\documentclass[]{ceurart}

\usepackage{listings}
\usepackage{lipsum}
\begin{document}

\copyrightyear{2024}
\copyrightclause{Copyright for this paper by its authors.
  Use permitted under Creative Commons License Attribution 4.0
  International (CC BY 4.0).}

\conference{IntRS’24: Joint Workshop on Interfaces and Human Decision Making for Recommender Systems, October 18, 2024, Bari (Italy)}

\title{Designing an Interpretable Interface for Contextual Bandits}

\author{Andrew Maher}[email=andrew@metica.com,]
\cormark[1]
\address{Metica, London}
\author{Matia Gobbo}[email=matia@metica.com,]
\author{Lancelot Lachartre}[email=lancelot@metica.com,]
\author{Subash Prabanantham}[email=subash@metica.com,]
\author{Rowan Swiers}[email=rowan@metica.com,]
\author{Puli Liyanagama}[email=puli@metica.com,]

\cortext[1]{Corresponding author.}

\begin{abstract}
Contextual bandits have become an increasingly popular solution for personalized recommender systems. Despite their growing use, the interpretability of these systems remains a significant challenge, particularly for the often non-expert operators tasked with ensuring their optimal performance. In this paper, we address this challenge by designing a new interface to explain to domain experts the underlying behaviour of a bandit. Central is a metric we term ``value gain'', a measure derived from off-policy evaluation to quantify the real-world impact of sub-components within a bandit. We conduct a qualitative user study to evaluate the effectiveness of our interface. Our findings suggest that by carefully balancing technical rigour with accessible presentation, it is possible to empower non-experts to manage complex machine learning systems. We conclude by outlining guiding principles that other researchers should consider when building similar such interfaces in future.
\end{abstract}


\maketitle

\section{Introduction}

Complex personalized recommender systems have become vital to building engaging, modern user experiences across a variety of domains \cite{10.1145/3589335.3648339,10.1145/3604915.3608873,Liu2023}. Although powerful, these systems cannot properly function without a human operator in place who can deploy and manage their correct running. These people -- typically non-experts in statistics and machine learning -- are expected to make reasoned, higher-order decisions about the recommender system, to maximize its performance and ensure it adds holistic value to the broader environment in which it sits. 

By default, however, modern recommender systems are complex and hard to interpret \cite{afchar2022explainability, steck2021deep}. They comprise multiple interlocking parts, each of which requires a strong mathematical background to understand. Take contextual bandits. They are an increasingly popular methodological approach that address known challenges such as the cold-start problem \cite{nguyen2014coldstartproblemsrecommendationsystems} and non-stationary environments \cite{hong2020nonstationarylatentbandits, li2020unifyingclusterednonstationarybandits}. Despite their efficacy as a recommender system method, providing a robust interpretation to their decisions is an unsolved problem. Similar to traditional supervised learning systems, they depend on black-box models to estimate the expected performance of recommendable items given a context. This difficulty is compounded by several factors: observational data only becomes available for the items selected by the bandit; interpretation is required not only for a single output but for multiple items from which the bandit is choosing; the ongoing modulation between exploration and exploitation means a bandit system does not always select the arm it predicts as most valuable.

For the non-expert human operator, several higher-order considerations need to be made to ensure each bandit-based recommender is continually well-tuned. Is it performing well enough to keep it in production? Should new arms be added, or existing ones removed? Are the context fields considered by the bandit sufficiently discriminatory to yield interesting results? Answering these questions requires an understanding of the underlying system that is both deep and broad. Yet there is a gap between this need for interpretation and the tools and interfaces that exist to provide it. 

A comparison with a sister decision-making domain is apt. In A/B testing, there is a well-known set of metrics and visualizations that determine which of the options being tested is best \cite{greenland2016statistical, biau2010p}. Statistical significance and MLE-based uplift charts predominate the field. Moreover, there exist dozens of commercially available platforms  with interfaces designed for easy interpretation \cite{fabijan2023b}, and probably thousands of proprietary in-house equivalents \cite{8819471, kaufman2017democratizingonlinecontrolledexperiments}. Few such equivalent interfaces exist for contextual bandits; none are publicly available.

To address these challenges, we have developed an intuitive interface designed to explain the behavior of a contextual bandit system. It is in production and being used in a commercial setting. By leveraging techniques from data visualization, off-policy evaluation and user-centric design, our interface aims to make the inner workings of a bandit system comprehensible to domain-expert operators. Central to the interface is a generic metric framework we term "value gain" -- a measure derived from off-policy evaluation that provides a clear indication of the real-world value of different elements of the system. 

It is important to clarify that our target audience is not the end-users who receive recommendations; significant research has already been conducted on boosting interpretability for these people \cite{afchar2022explainability, tsai2019evaluating, musto2019combining, gao2023chat, tan2021counterfactual}. We focus instead on the people who choose the inputs to the recommender system, and who are responsible for its proper functioning. We assume nothing about their knowledge of statistics, only that they can read a quantitative dashboard. As well, it is worth noting that although the interface has been designed for contextual bandits, and works with any underlying bandit algorithm, the ideas apply equally well to any recommender methodology. The sole requirement is that the system comprise a (relatively) limited number of options from which to choose -- and that there is value in understanding their respective performance. We do not try to solve for the problem of choosing from an ever-changing and very large library of options (as in, say, a video recommender system).

The rest of the paper is organised as follows. In Section \ref{Related work} we discuss related work. We then present our developed interface in Section \ref{Interface}, followed by a user study in Section \ref{User study} to evaluate its effectiveness. Finally, in Section \ref{Conclusions} we outline guiding principles for future practitioners looking to build similar dashboards, then discuss future directions in this space.

\section{Related work} \label{Related work}

\emph{Contextual Bandits} can be viewed as an extension of traditional experimentation in which the arm-assignment decision is both automated (hence bandit) and personalized (hence contextual). Typically, they comprise a reward model and a policy. The former governs the bandit's understanding of the world, with common choices including linear regression \cite{chu2011contextual} and neural networks \cite{pmlr-v139-nabati21a}. The latter dictates how the bandit modulates between exploration and exploitation. Canonical examples are UCB (Upper Confidence Bound) \cite{abbasi2011improved}, and Thompson sampling \cite{agrawal2013thompson}, with numerous applications across various domains such as online advertising \cite{han2020contextual}, personalized news feeds \cite{li2010contextual}, customer support \cite{sajeev2021contextual}, and e-commerce recommendations \cite{hill2017efficient, 10.1145/3383313.3412234}.

\emph{Off-Policy Evaluation} is the main paradigm through which the efficacy of different bandit approaches is evaluated. It is a counterfactual estimation procedure in which the logged policy -- the one for which real-world data is observed and measured -- is compared to a hypothetical target policy. Numerous estimators exist to facilitate this comparison, including inverse propensity scoring, the direct method, and doubly robust estimators \cite{farajtabar2018robust, dudik2011doubly, wang2017optimal}. These techniques allow for the assessment of new policies without the need for costly and time-consuming online experimentation.

In contrast to the "how good," \emph{Explainable AI} attempts to elucidate the purer "how" and "why" of machine learning approaches \cite{mohseni2021multidisciplinary}. Recent research in this field has focused on identifying new methods, different presentation approaches, and better ways to judge the goodness of these explanations. The vast majority focus on interpretability concerns with respect to the recipient of a recommendation, something that is out of the scope of this work. Of particular relevance are a number of \emph{User Interfaces} designed to enable the proper understanding of different ML systems. For example, ActiVis and LSTMVis are two different visualisation interfaces for interpreting deep learning models and results \cite{8022871, strobelt2017lstmvis}, and InterpetML is a holistic system to help understand ensembles of decision trees \cite{nori2019interpretml}. A number of similar explanations and interfaces exist for more general reinforcement learning policies \cite{mishra2022not, milani2024explainable}. Although many organisations deploy contextual bandits -- and some offer them as services to other companies -- we could not find any equivalent bandit interfaces in the literature.

\section{Interface} \label{Interface}

\begin{figure}[htbp]
  \centering
  \includegraphics[width=\linewidth]{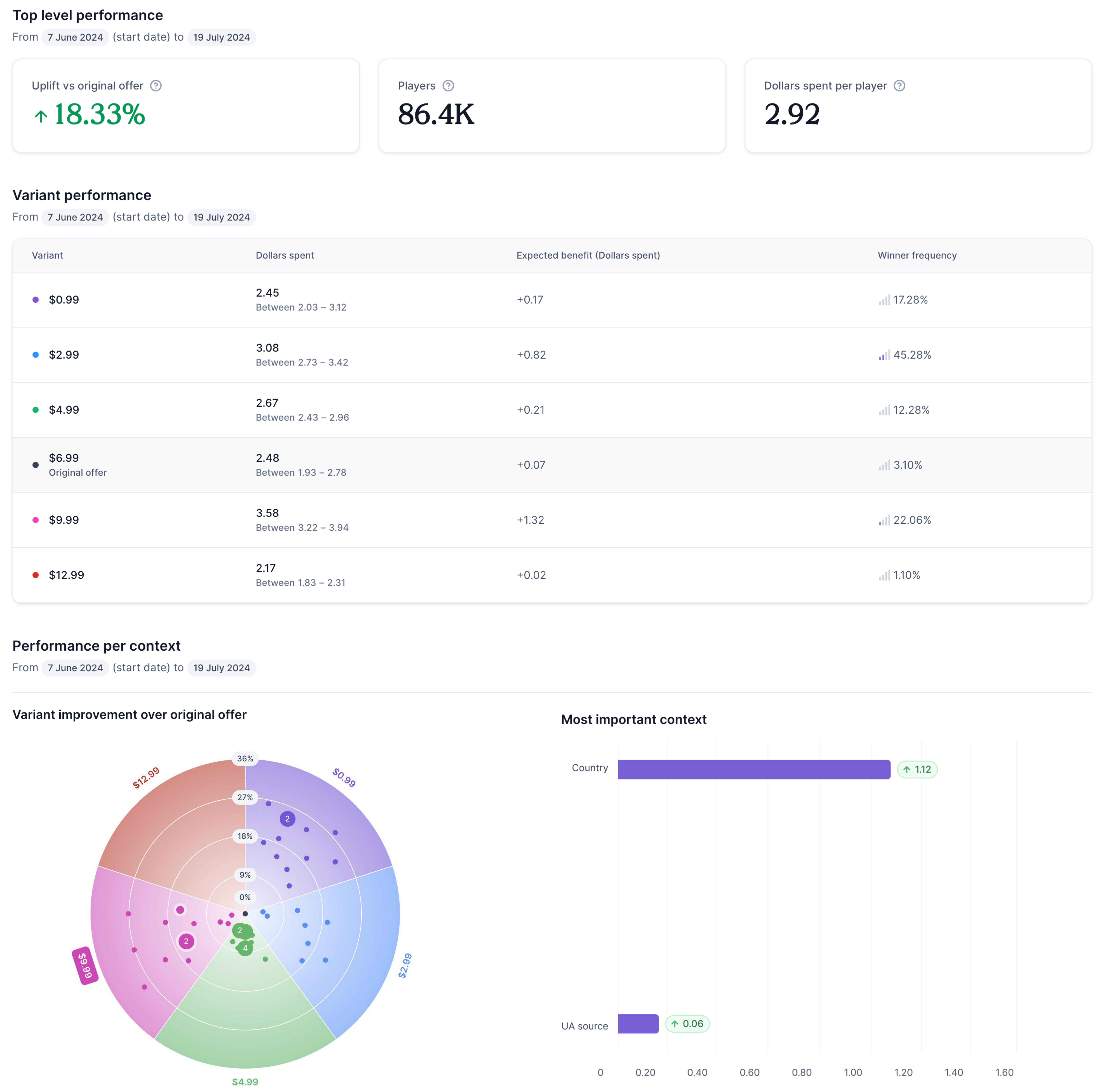}
  \caption{User interface for contextual bandits. The interface comprises three main components: top level performance, variance performance and performance per context. They each describe different elements of the performance of the bandit system, in increasing granularity of units.}
  \label{fig:User interface}
\end{figure}

Figure \ref{fig:User interface} shows the interface we designed, through which human operators can interpret the workings of a contextual bandit system. It contains multiple, ordered visualizations that aim to provide increasing detail about the performance of different aspects of the bandit.

The audience of the interface are the people responsible for launching -- and potentially altering -- the bandit. They need to understand not just its holistic performance, but also how each component of the bandit contributed to that performance. One way of providing this understanding is through comparison: supposing that component wasn't included in the system, how much less value would the bandit generate? In other words, what is the value \textit{gained} by the inclusion of that component.

To this end, we introduce the \emph{value gain} metric -- an estimate of the value of the production bandit, with respect to a simpler one in which certain components are ablated. Below, we define this metric in its general form. We then describe the interface itself in detail -- considering as a prototypical example a use-case in which a mobile game wishes to increase dollars spent on in-app purchases. Although we consider this example for the purposes of the paper, the interface applies equally well to any use-case served by a contextual bandit.

\subsection{Value gain} \label{Value gain}

Let $r$ denote the reward observed by the contextual bandit from one of its actions, and let $\pi$ be the combination of policy-and-context-model that chooses actions yielding $r$. We can define the \emph{value of $\pi$} as
\begin{equation}
    v^{\pi} = \mathbb{E}_{\pi}\left[ r \right].
\end{equation}
Here $v$ is measured in the same units as the optimisation goal of the bandit. It describes, for example, the average revenue per user achieved by the bandit.

This policy-and-context-model $\pi$ comprises multiple elements: the choice of exploration agent; the fields introduced to represent user context; the arms selected for inclusion in the bandit. Each of these individual components contributes to the value of the overall system. We attempt to measure that contribution through ablation. This is the principle behind the value gain metric.

Suppose we had an alternative policy-and-context-model $\overline{\pi}$. And further suppose that $\overline{\pi}$ is equivalent to $\pi$ in all but some elements (which we'll denote as $\tau$) -- ie, its components are a subset of those present in $\pi$. We define the \emph{value gain} -- the gain in value produced by those missing elements -- as being given by
\begin{equation}
    g(\tau) = v^{\pi} - v^{\overline{\pi}} = \mathbb{E}_{\pi}\left[ r \right] - \mathbb{E}_{\overline{\pi}}\left[ r \right].
\end{equation}

It is impractical to calculate online $v^{\overline{\pi}}$ across the range of elements in which we might be interested. Doing so would require diverting a large proportion of traffic to alternative assignment algorithms, reducing the volume of data from which the main policy can learn and exposing multiple users to a potentially inferior bandit. Instead, we use methods introduced in the off-policy evaluation literature \cite{wang2017optimal, dudik2011doubly}. Although these methods are not perfect, they nonetheless represent the gold standard by which bandit policies are evaluated offline.

In practice any off-policy estimator can be used. To make concrete the ideas within this paper, we consider the inverse-propensity score estimator \cite{dudik2011doubly}. It estimates $v^{\overline{\pi}}$ as
\begin{equation} \label{eq:Value gain}
    v^{\overline{\pi}} = \frac{1}{n}\sum_{i=1}^{n} \frac{\overline{\pi}(a | x_i)}{\pi(a | x_i)} r(x_i),
\end{equation}
where the functions in the fraction represent the probability of assigning arm $a$ to context $x_i$ for the ablated and actual policy respectively.

For example: suppose an e-commerce website launched a bandit to improve conversion rate on its landing page, by launching different variants of its home page. And suppose further that one of the variants is the pre-existing home page, ie, the one that existed before the launch of the bandit. Here, $g(\tau)$ can be used to measure the value gained from including all non-baseline arms. In this case, $\tau$ denotes the set of non-baseline arms, $\overline{\pi}$ is a trivial policy that assigns the baseline homepage to all incoming traffic, and $\pi$ is the contextual bandit launched by the e-commerce website. The value gain $g(\tau)$ shows the extra conversion rate introduced by the bandit with respect to the original baseline experience.

\subsection{User interface}

The interface in Figure \ref{fig:User interface} comprises three main sections:
\begin{enumerate}
    \item \textbf{Top level performance}: Short summary of the bandit's overall performance. It provides an overview that can be digested in as short a time as possible.
    \item \textbf{Variant performance}: Description of how each arm in the bandit in the bandit is performing. It enables an operator to determine which arms performs best, and which are candidates for removal from the system.
    \item \textbf{Performance per context}: More granular information about the relationship between expected performance and the different contexts used within the bandit. It provides further understanding of the way in which the bandit is personalising the underlying experience.
\end{enumerate}

\subsubsection{Top level performance}

The top level performance section presents three distinct metrics:
\begin{enumerate}
    \item \textit{Uplift vs original offer}: What is the percentage increase provided by the bandit on the goal metric, when compared to a baseline arm?
    \item \textit{Players}: How many people have been exposed to the bandit so far?
    \item \textit{Dollars spent per player}: What is the performance of the bandit on its revenue-maximising goal metric so far?
\end{enumerate}

The metrics are explicitly presented in order of importance. First, how much value is the bandit adding its use-case? Second, is it reaching a sufficiently large population? Third, what is its average performance?

The first metric is based off the previously defined \emph{value gain}. Here, we assume that one of the arms can be considered as a ``baseline'' -- it is the arm that would be shown to all traffic, if the recommendation policy was not in use. In this case, we show how the policy performs relative to another one that contains only that baseline arm. Following the notation above -- and assuming the production policy comprises the set of arms $a_k \in \mathcal{A}$, we calculate $g(\tau)$ for
\begin{equation}
    \tau = \left\{a_k \in \mathcal{A}: a_k \neq a_{\text{baseline}} \right\},
\end{equation}
ie, the counterfactual policy contains only the baseline arm $a_{\text{baseline}}$.

\textit{Players} and \textit{Dollars spent per player} are calculated directly from logged assignment and rewards data.

\subsubsection{Variant performance}

The variant performance table shows arm-level information. It is a table comprising one row per arm and four columns. One of the columns describes the arm itself (with a name or other distinguishing information), the other three  summarise performance information about that arm. 

The first metric column -- \textit{Dollars spent} -- shows both (A) the expected performance of the arm for all people exposed to the bandit, as well as (B) a range of potential performance values. We calculate the expectation and range by first estimating for a given its reward across all players. Then, on that distribution of estimates, we compute three summary statistics: the mean, and the 10th and 90th percentiles.

The second column is \textit{Expected benefit (Dollars spent)}. It measures the achieved value that can be attributed to the arm in question. We again use \emph{value gain} to evaluate this quantity. In this instance, we compare the production policy to a counterfactual one whose ablated element is $\tau = a_k$, where $a_k$ is the arm being measured. Take variant \$0.99 as an example. It has an expected benefit of +0.17. This means the bandit gains 17 cents more per user thanks to its ability to show this arm.

The third column shows the proportion of contexts for which that arm was displayed to users.

\subsubsection{Performance per context}

The performance per context component contains two visualizations. The first is a radar chart in which a circle is split into multiple segments, with each segment representing a single arm. Dots are plotted on to the segments. Each dot represents a distinct context vector encountered by the bandit. The dots are placed into the segment corresponding to the expected best arm for that context vector. Their distance from the chart's origin is defined by the relative uplift of that arm compared to the original offer. We again use the \emph{value gain} to calculate this distance.

The second visualisation is a bar chart shows the value gain attributable to each context field. Here we compare the production policy to a counterfactual one, in which the context field in question is removed. Each bar hence describes how much better is the bandit thanks to the inclusion of that context field.

\section{User study} \label{User study}

To better understand the ability of our interface to meaningfully represent interpretable results from a contextual bandit system, we conducted a user study. 

\begin{table}[!ht]
    \caption{Results from the self-guided component of the user study.}
    \label{table:self-guided-results}
    \centering
    \begin{tabular}{>{\raggedright}m{0.2\textwidth}m{0.7\textwidth}} 
        \toprule
        Introduction to interview and page & All interviewees quickly discerned the meaning of the top-level performance metrics and how they would help in measuring performance. \\ \midrule
        Variant performance table & The variant performance table was the second element of the page at which they each arrived. All three understood the dollars spent and winner frequency columns, but needed some prompting with the latter. 
        
        More difficult was the expected benefit column. Each interviewee correctly stated it denoted the value of the variant, and that the measure was comparing the variant to something else. No interviewee could state what that something else was. At first, each said it might be the baseline variant before deducing that to be impossible (as the baseline variant also had a non-zero value). Even after extensive prompting, they couldn't correctly define the metric. \\ \midrule
        Radar chart & The radar chart was the last component of the page each interviewee discovered. They all found it somewhat daunting to explore at first, but quickly established (A) what each point represented and that (B) each segment related to an individual variant. Two of the interviewees stated the correct definition of the dots' placement. All three worked out how to evaluate the different variants using the chart. One candidate noted the chart was pretty but potentially superficial. \\ \midrule
        Desire for contextualisation & Beyond the specific sub-components of the page, the three interviewees each expressed a desire for more context beyond the base numbers shown. All three explicitly requested information on the ``significance'' of the results. Two wanted to understand the number of observations relating to each number. One interviewee requested filters to gain more granular information about the data. \\
        \bottomrule
    \end{tabular}
\end{table}

\begin{table}[]
    \centering
    \caption{Results from the knowledge-based questions in the user study.}
    \label{table:knowledge-based-results}
    \begin{tabular}{>{\raggedright}m{0.2\textwidth}m{0.7\textwidth}}
        \toprule
        Bandit value & All interviewees correctly stated the bandit provided value above the baseline. They relied only on the top-left uplift metric to make this point. When asked whether they'd let the bandit continue running, all three replied yes. They again depended on the top-left uplift metric. \\ \midrule
        Variant performance & All three interviewees could reason about which variants were worse-performing. They used a combination of the winner frequency, the radar chart and the expected benefit to answer -- with no clear preference among these elements. They all determined a best set of variants (\$2.99 and \$9.99) using the same elements. None tried to contrast the quality of these two variants (using the expected benefit column and winner frequency, for example). \\ \midrule
        Context contribution &  Two interviewees used the context contribution chart to reason that removing poor-performing context fields would improve the bandit (by avoiding opportunity cost and/or simplifying the system). One interviewee couldn't reason effectively about which contexts best contributed to the bandit. They instead lent on their practical experience (of user behaviour in different countries). They didn't try to use the context contribution chart for their answer. \\ 
        \bottomrule
    \end{tabular}
\end{table}

\subsection{Format}
As in \cite{8022871} and \cite{purificato2021evaluating}, we performed a qualitative evaluation built off deep-dive interviews with candidates who would use this system as part of their daily workload. The interviews each took forty-five minutes. We started with a short explanation of how contextual bandits work, to ensure candidates had sufficient familiarity with the topic. Interviewees were then encouraged to explore the UI on their own and, whenever they focused on a specific component, they were asked to explain their perception of what it meant.

During the self-guided exploration, we also asked each interviewee the following specific knowledge-based questions to probe the extent of their ability to correctly interpret the bandit using the interface:
\begin{itemize}
    \item \textbf{Bandit value:} How is the bandit performing compared to a default experience? When do you think the optimization would/should stop?
    \item \textbf{Variant performance:} What are the best / worst performing variants? Why? Given the information presented, would you intervene to change anything about the variants? If so, what changes would you make?
    \item \textbf{Context contribution:} Given the information presented, would you intervene to change anything about the context fields being used? If so, what changes would you make?
\end{itemize}

\subsection{Results} \label{Results}

We conducted three such deep-dive interviews in total. Each interviewee was a marketing professional who would be the person responsible for interpreting bandit choices and outcomes to make operational decisions. All of them had extensive experience with A/B testing, but little practical background usage of a contextual bandit. Each interviewee was shown the interface as depicted in Figure \ref{fig:User interface}. Results from the self-guided exploration and knowledge-based questions are summarised in Tables \ref{table:self-guided-results} and \ref{table:knowledge-based-results}.

\section{Conclusions} \label{Conclusions}

\subsection{Guiding principles}

We have presented a visual interface to explain the workings of an in-operation contextual bandit system, built using novel metrics underpinned by methods from off-policy evaluation.

Through this exercise, we can identify a number of broad, guiding principles to inform the useful design of a similar interfaces in future. These principles are outlined in Table \ref{table:principles}. Future researchers and practitioners should use them to help direct their own design processes. 

Of these, the two most crucial are the complementary pair: \textit{Feel empowered to use technical tools} / \textit{Use clear non-technical language}. In the context of machine learning and recommender systems, the most insightful metrics can be simple to understand but inherently complicated to calculate. If they add the most value -- use them. For example, we introduced ideas from off-policy evaluation to our dashboard. Despite their relevance, off-policy evaluation is not particularly well-known outside the machine learning and statistics community. As we built the interface, we held concerns internally that our audience might feel uncomfortable with these metrics, and not trust them sufficiently. However none of our interviewees raised a problem.

To caveat the above point, its important to consider the audience that will read the results. Think about the implication the statistical machinery conveys, then use that as the description of the metric. Don't blindly use the jargon term that aligns best with the literature if an end-user will not understand its meaning -- an error we made in our design.

\begin{table}[]
    \centering
    \caption{Guiding principles for an interpretable interface for a contextual bandit. Future researchers and practioners should consider these when building their own, similar interfaces.}
    \begin{tabular}{>{\raggedright}p{0.2\textwidth}p{0.7\textwidth}}
        \toprule
        Principle & Explanation \\
        \midrule
        Feel empowered to use technical tools  & Sometimes the most relevant metrics are highly technical. Don't shy from their usage. We used techniques from off-policy evaluation, an alien field to our interviewees. No one raised a concern; our candidates trusted and accepted the information we shared. \\ \midrule
        Use clear, non-technical language & Describe results in a way the audience can easily reason about. We named one column ``expected benefit'', as it related to statistical expectation and indirectly conveyed meaning. The title made sense to us as statisticians; our interviewees didn't get it. By contrast, all candidates understood ``uplift vs original''. \\ \midrule
        Consciously order information & Different results exist within the hierarchy of complexity in a recommender system. Carefully consider what to show and when. Our interviewees could digest our most complex visualisation -- an information-dense radar chart -- precisely because they'd been carefully shown simpler results earlier on.  \\ \midrule
        Contextualise results & A repeated criticism of our interface was a lack of statistical significance or volume information. People couldn't reason about the importance of the results they were seeing. Providing contextual information of this sort enables operators to respond proportionately to insights. \\ \midrule
        Facilitate decision-making & Fundamentally, insights are useful only if they lead to decision-making. Consider what will guide choices, and present it in complementary formats where useful. Our interviewees successfully answered our task questions by combining multiple elements of our interface. It was this action-oriented approach they praised most. \\
        \bottomrule
    \end{tabular}
    \label{table:principles}
\end{table}

\subsection{Future work} \label{section:future-work}

In this paper, we explore ideas and designs to enable the useful interpretation of a single contextual bandit system containing a (relatively) limited number of meaningful options. Relaxing these two constraints -- interpreting only a single bandit, and considering a much wider set of options -- would require additional design considerations. 

Take multiple bandits: our explorations here relate to providing a deep-dive on a narrow, single optimisation. This necessitates an abundance of information that becomes hard to parse when multiplied across use-cases. This is what would be found in practice -- one bandit managing a search experience, another bandit the delivery of product details, and a third bandit the creatives to display. We've so far considered an initial solution of surfacing a single key metric for each bandit (the \textit{uplift vs original offer} metric). Future research could improve on this by (A) more robustly determining which metric is the most salient to display, (B) ascertaining how to usefully triage among running bandits and (C) explaining the health of all systems in parallel.

Considering a much wider number of arms is another interesting technical challenge. Components we've introduced here -- the radar chart, the table of performance per variant -- do not naturally extend to the case where there is more than, say, twenty options. But modern contextual bandit systems, particularly with the advent of generative AI, can be easily designed to have much larger numbers of meaningful variants. Conveying information from across a multitude of potentially differing variants is something we hope to consider in future work.

\bibliography{references}

\end{document}